%% file: paper.tex
\pdfoutput=1
\documentclass[runningheads]{llncs}
\usepackage[T1]{fontenc}
\usepackage{todonotes}
\usepackage{wrapfig}
\usepackage{subfigure}
\usepackage{amsmath,amssymb,amsfonts}
\usepackage{algorithmic}
\usepackage{graphicx}
\usepackage{textcomp}
\usepackage{subfigure}
\usepackage{xcolor}
\usepackage{tikz}
\usepackage{paralist}
\usepackage{enumitem}
\usepackage{calc}
\usetikzlibrary{shapes.geometric, arrows, positioning}
\usepackage{tabularx}
\usepackage{booktabs}
\usepackage{listings}
\usepackage{color,soul}
\usepackage{hyperref}

\newcolumntype{P}[1]{>{\centering\arraybackslash}p{#1}}

\begin{document}
\title{Automated Business Process Analysis: An LLM-Based Approach to Value Assessment \\ \lbrack Extended version\rbrack}
\titlerunning{Automated Business Process Analysis}
%
\author{William~De~Michele \and
Abel Armas Cervantes \and
Lea Frermann}
\authorrunning{William~De~Michele et al.}
%
\institute{The University of Melbourne, Australia \\
wdemichele@student.unimelb.edu.au\\
\{abel.armas,lea.frermann\}@unimelb.edu.au}
\maketitle              
\begin{abstract}
Business processes are fundamental to organizational operations, yet their optimization remains challenging due to the time-consuming nature of manual process analysis. 
Our paper harnesses Large Language Models (LLMs) to automate value-added analysis, a qualitative process analysis technique that aims to identify steps in the process that do not deliver value. To date, this technique is predominantly manual, time-consuming, and subjective. 
Our method offers a more principled approach which operates in two phases: first, decomposing high-level activities into detailed steps to enable granular analysis, and second, performing a value-added analysis to classify each step according to Lean principles. This approach enables systematic identification of waste while maintaining the semantic understanding necessary for qualitative analysis. 
We develop our approach using 50 business process models, for which we collect and publish manual ground-truth labels. Our evaluation, comparing zero-shot baselines with more structured prompts reveals
\begin{inparaenum}[(a)]
\item a consistent benefit of structured prompting and 
\item promising performance for both tasks. 
\end{inparaenum}
We discuss the potential for LLMs to augment human expertise in qualitative process analysis while reducing the time and subjectivity inherent in manual approaches.\footnote{Code, data, prompts are made available at \url{https://github.com/wdemichele/WIPE}.}

\textbf{Keywords:} Business Process Management, Waste Identification, Large Language Models, Value-Added Analysis, Process Analysis
\end{abstract}
\input{tex/introduction}
\input{tex/relatedwork}
\input{tex/wipe}

\input{tex/evaluation}

\input{tex/conclusion}

\bibliographystyle{plain}
\bibliography{bibliography}

\end{document}

%% file: tex/introduction.tex
\section{Introduction}

Businesses are under constant pressure to optimise their operations, reduce costs, and enhance customer satisfaction \cite{waddock2002responsibility}. Business Process Management (BPM) is a discipline that enables organisations to model, analyse, and improve their operational processes~\cite{lee1998business,dumas2018fundamentals}. A key phase in the BPM lifecycle is process analysis, where inefficiencies and problematic areas are identified in the as-is process to inform process redesign. Central to this analysis is the identification of waste -- activities or steps in the process that do not add value from the customer's perspective or are unnecessary for the business to function effectively~\cite{mansar2007best}. 

Traditional approaches to waste identification rely on manual expert analysis, which can be time-consuming, costly, and prone to subjectivity \cite{rudden2007making}. As organisations scale and processes become more complex, there is a growing need for automated and systematic methodologies that can efficiently analyse business processes and accurately identify areas of waste. {\it Value-added analysis} is a qualitative analysis technique that consists of identifying activities or steps in the process that do not deliver value to the customer nor to the business, and which can be potentially removed during process redesign. 


\begin{figure}[t]
    \centering
    \includegraphics[width=\textwidth]{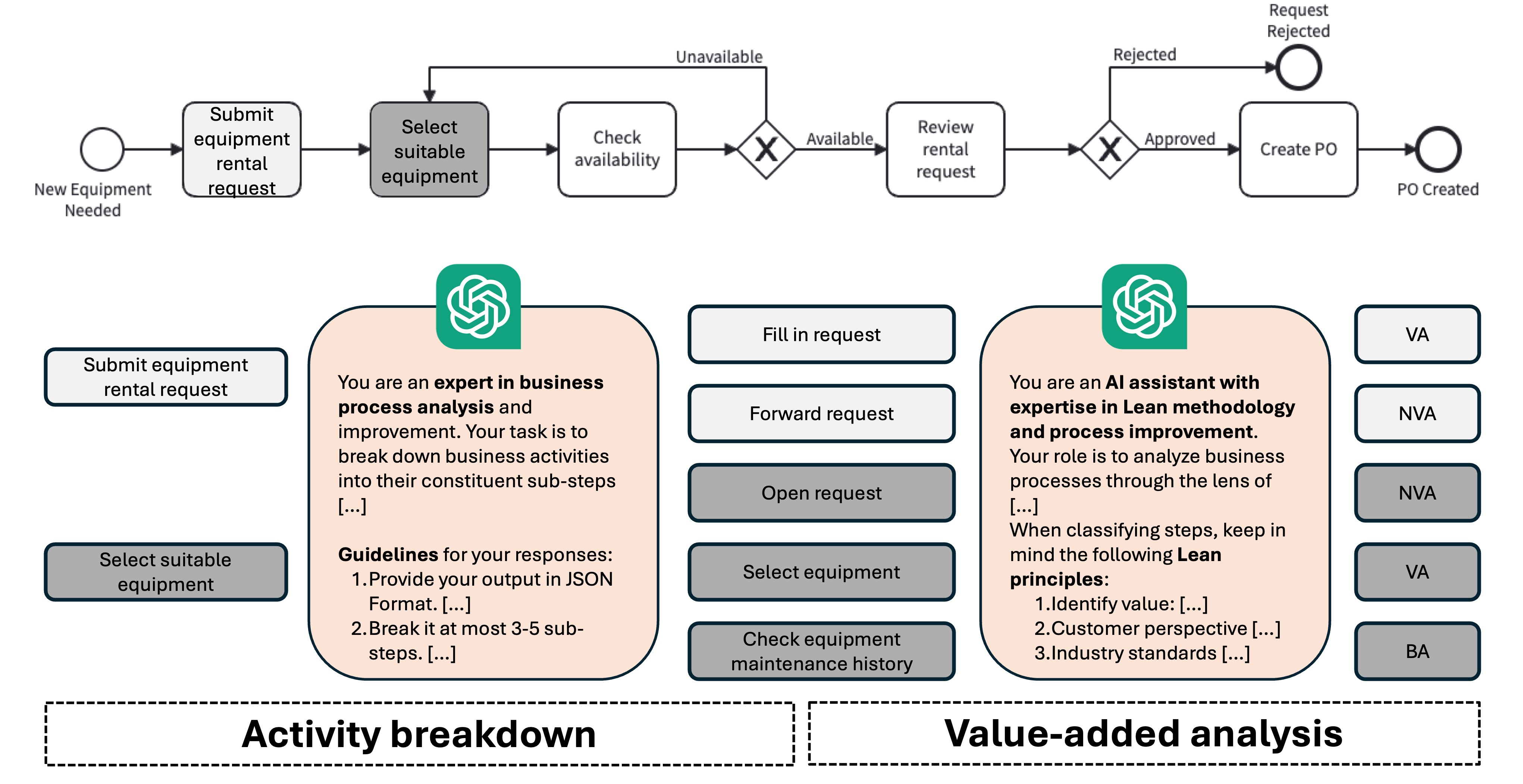}
    \caption{Overview of our LLM-based value-added analysis illustrated on an excerpt of a rental equipment process (top;  \cite{dumas2018fundamentals}). The first two activities (left) are broken down into multiple steps (center), which are then labeled as value adding (VA), business value adding (BVA) or non-value adding (NVA) (right).}
    \label{fig:pipeline}
\end{figure}

This paper introduces a novel approach that leverages Large Language Models (LLMs) to automate the identification of unnecessary steps in business processes (Fig.~\ref{fig:pipeline}). 
LLMs have demonstrated remarkable capabilities in natural language understanding and generation~\cite{allen1988natural,karanikolas2023large}, making them suitable for semantic analysis of business processes~\cite{berti2024pm,MendlingBachhofnerVigdof2023}. Our approach operates in two phases: first, high-level activities are decomposed into fine-grained steps, and second, a value-added analysis~\cite{eakin2020value} is performed over the steps to classify them as Value Adding (VA), Business Value Adding (BVA), or Non-Value Adding (NVA) based on Lean principles~\cite{womack1997lean}. 
To evaluate our approach, we created a dataset of 50 business processes with BPMN models spanning various industries, including banking, healthcare, customer service, and technology. We manually annotated these processes with activity-to-step breakdown, and step-level value-added analysis. Although we presented our approach using process models as a starting point, the approach can be applied to event logs by focusing on the activities represented.

The remainder of this paper is organized as follows: Section~\ref{sec:literature_review} reviews related work on LLMs in business process management, Section~\ref{sec:methodology} presents our methodology, Section~\ref{subsec:experimental_setup} details our evaluation, Section~\ref{sec:discussion} discusses the implications of our findings, and Section~\ref{sec:conclusion} presents conclusions and future work.

%% file: tex/relatedwork.tex
\section{Related Work}
\label{sec:literature_review}

The natural language generation capabilities of recent large language models (LLMs) offer new opportunities for automating complex tasks, including in the field of business process management~\cite{MendlingBachhofnerVigdof2023}. For example, LLMs have been used for discovering process models from text, or vice-versa~\cite{berti2024pm,berti2023abstractions,GrohsAER2024}. In this section we review relevant research on the application of LLMs for business process analysis. 

\subsubsection{Large Language Models and Prompt Engineering.}

LLMs combine transformer architectures with attention mechanisms \cite{vaswani2017attention} with self-supervised pre-training on massive text corpora followed by instruction tuning. This optimises the models' abilities to respond to natural language inputs (prompts). This strategy has been shown to lead to remarkable capabilities in natural language understanding, generation, and reasoning. 

The quality of LLM responses largely depends on prompt engineering \cite{marvin2023prompt} -- the practice of designing input instructions that elicit desired model behaviors. Techniques range from simple prompting (zero-shot; \cite{xian2017zero}) over extending prompts with input-output examples (few-shot; \cite{wang2020generalizing}), to more sophisticated approaches like chain-of-thought reasoning, which guides models through step-by-step reasoning processes~\cite{wei2022chain}. Structured prompting techniques \cite{jiang2023effective} incorporating role descriptions \cite{kong2024self}, task specifications, and output formats have proven effective for specialised tasks \cite{zhang2023instruction}. Recent research has explored systematic prompt optimization through automatic generation and iterative refinement, with the aim of identifying optimal prompt structures for specific applications \cite{zhu2023prompt,khattab2023dspy}. These advanced prompting techniques form the foundation for our approach to activity breakdown and value-added analysis.

\subsubsection{Waste Analysis with LLMs.}

Lashkevich et al.~\cite{lashkevich2024llm} present an LLM-based approach for the analysis and redesign of business processes to optimize waiting time. The approach uses five types of waiting time that can be identified on an event log~\cite{LASHKEVICH2024102434}-- batching, resource contention, prioritisation, resource unavailability, and extraneous factors. The approach combines process mining outputs with structured prompting to generate both technical and strategic recommendations for process redesign. It is shown structured knowledge bases significantly outperform unstructured documentation for LLM prompting, with their enhanced method achieving 68\% pattern utilization compared to 15\% for unstructured approaches. Their findings also reveal a key insight about LLM-based process analysis: different prompting approaches serve distinct purposes, with enhanced prompting generating specific, actionable recommendations for analysts, while baseline prompting produces strategic insights for management decision-making.
While~\cite{lashkevich2024llm} demonstrates the potential of combining LLMs with process mining for waste identification through temporal analysis, they address only one aspect of process waste. Our work on Activity Breakdown and Value-added Analysis complements these efforts by focusing on qualitative waste identification through semantic understanding of activities. The combination of our approach with others, such as~\cite{lashkevich2024llm}, enables a more comprehensive understanding of process waste and improvement opportunities.


While quantitative analysis has dominated much of the existing research of LLM applications in BPM (e.g.,~\cite{berti2023abstractions,lashkevich2024llm,JessenSrokaFahland2023}), focusing on metrics such as process performance, execution time, and resource utilisation, there has been a notable lack of emphasis on qualitative aspects of processes. 
Our research addresses this gap by incorporating qualitative, value-added analysis approaches similar to those proposed by~\cite{eakin2020value}. By leveraging LLMs to perform in-depth qualitative analysis of process activities, we provide insights into the semantic and contextual aspects of business processes that are often overlooked in traditional quantitative approaches. This qualitative focus allows for a more nuanced understanding of process value and waste, considering factors such as the perceived importance of activities to stakeholders, alignment with business perspective, and complex contextual activity relationships - aspects that are challenging to capture through purely quantitative means. 
The approach presented in this paper demonstrates the practical applicability of LLMs in enhancing qualitative aspects of business process management, complementing existing quantitative methods, and providing a more complete picture of process efficiency opportunities.

%% file: tex/wipe.tex
\section{Automated Value-Added Analysis}
\label{sec:methodology}

This section presents the main contribution of the paper, an automated approach to value-added analysis. It is divided into two stages. First, {\bf Activity Breakdown} divides tasks into individual steps and, second, {\bf Value-Added Analysis} categorizes each step with its perceived value.
Each of the subsections below details the development of prompt components \cite{sahoo2024systematic} and prompt optimisation. Effective prompt engineering is critical for guiding LLMs to produce accurate and relevant outputs~\cite{marvin2023prompt,giray2023prompt}. The design of prompts influences the LLM's understanding of the task, as well as the quality of the generated content. For each stage, we include a simple \emph{zero-shot} prompting approach as a baseline. Additionally, based on our domain knowledge and best-practices in LLM prompting, we devise a series of \emph{Structured Prompts} for each task (role ascription, guidelines and examples) and identify the component values that lead to optimal performance. 

\subsection{Activity Breakdown}
\label{subsec:activity_breakdown}

Business processes models can represent activities with varying levels of granularity, which hinders an even comparison of the added value across tasks. Hence, to obtain comparable levels of granularity, activities are broken down into atomic ``steps'' (see Fig.~\ref{fig:pipeline} (left) for an example), which will be assessed during the value added analysis. 
We note that breaking an activity into steps can be subjective, as there can be various valid options (see Section~\ref{subsec:activity_breakdown_results} for further discussion). 

We develop a series of automated approaches to break activites into steps. A \emph{zero-shot baseline} is used as a control to assess the LLM's out-of-the-box capabilities without extensive guidance. The prompt included a concise instruction to break the activity into steps, without any guidelines, or examples. To improve on the zero-shot baseline, we established standardised \emph{Decomposition rules} to mitigate the inherent subjectivity. In addition, we design \emph{Structured Prompts} and systematically test the impact of different components on LLM performance. Our components are summarized in Table~\ref{tab:activity_breakdown}, and they cover 
\begin{inparaenum} [(1)]
\item the role assigned to the LLM,
\item various levels of detail in the task description,
\item various levels of guidelines,
\item the focus shift to analyse the steps, 
\item different numbers and kinds of examples included in the prompt~\cite{reynolds2021prompt}, and 
\item additional information.\footnote{Descriptions of all roles and the component values, for both activity breakdown and value-added analysis, can be found in our repository: \url{https://github.com/wdemichele/WIPE}.}
\end{inparaenum}

\begin{table}[t]
\footnotesize
\caption{Structured Prompt Components for activity breakdown. The optimal choice for each component is highlighted in bold.}
\label{tab:activity_breakdown}
\begin{tabularx}{\columnwidth}{lX}
\toprule
\textbf{Component} & \textbf{Variations} \\
\midrule[\heavyrulewidth]
Role Description & Neutral Analyst, Subject Matter Expert (SME), SME Detailed, \textbf{Business Process Expert}, Project Manager, Process Analyst, Operations Manager \\
\midrule[\lightrulewidth]
Task Description & Breakdown Substeps, \textbf{Breakdown with Dependencies}, Breakdown Focusing on Outcomes \\
\midrule[\lightrulewidth]
Guidelines & Standard Guidelines, \textbf{Detailed Guidelines}, Outcome-Focused Guidelines \\
\midrule[\lightrulewidth]
Focus Shift & Action-Focused, Outcome-Focused, \textbf{Process-Focused} \\
\midrule[\lightrulewidth]
Example Outputs & Zero-Shot (no examples), One-Shot (one example), \textbf{Few-Shot (multiple examples)}, Detailed One-Shot, Detailed Few-Shot \\
\midrule[\lightrulewidth]
Context & \textbf{Include Business Context}, Emphasise Order and Dependencies \\
\bottomrule
\end{tabularx}
\end{table}

\begin{figure}
    \centering
    \includegraphics[width=1\textwidth]{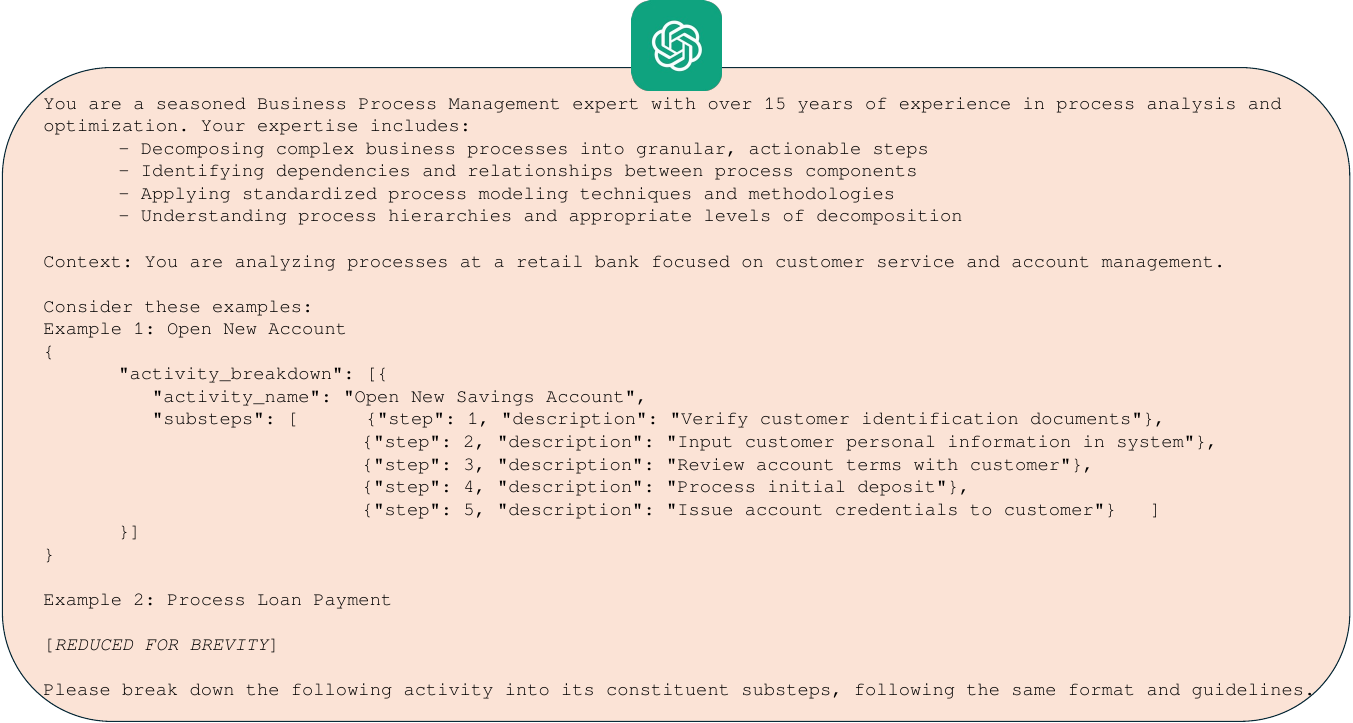}
    \caption{The final prompt used to obtain an activity breakdown from our LLM.}
    \label{fig:prompt_activity_breakdown}
\end{figure}

We obtained the optimal combination of components through greedy grid-search over the combinatorial space~\cite{liashchynskyi2019grid,wilt2010comparison} by systematically testing combinations of prompt components, selecting the best-performing component at each task while keeping previously selected components fixed. To do so, we first fixed each component to its baseline configuration, and then iteratively optimized each component while keeping all others fixed. Afterwards, we revisited components and testing additional prompt variations particularly if initial selections did not lead to significant improvements. Table~\ref{tab:activity_breakdown} shows the optimal prompt component values in bold, and Figure~\ref{fig:prompt_activity_breakdown} illustrates the full prompt.

\subsection{Value-Added Analysis}
\label{subsec:value_adding_analysis}

Value-added analysis \cite{eakin2020value} assigns a {\it value} to each step identified during the task breakdown. The value is a label that determines its contribution to the overall business process from a given perspective. Each step is assigned one of three possible labels:
\begin{itemize}
\item {\bf Value Adding (VA):} Steps that directly contribute to fulfilling customer needs and for which customers are willing to pay for;
\item {\bf Business Value Adding (BVA):} Steps that do not add direct value for the customer, but are necessary for business operations, such as for compliance or quality control; 
\item {\bf Non-Value Adding (NVA):} Steps that neither add value to the customer nor to the business, this includes unnecessary or redundant activities.
\end{itemize}

Given a business process broken down into a series of steps, we prompt the LLMs to 
\begin{inparaenum}[(a)] 
    \item assign a value label to each step and 
    \item provide a justification for each label, providing a rationale, enhancing transparency and interpretability. 
\end{inparaenum}
Figure~\ref{fig:pipeline} (right) shows an example of the classification applied to the step decomposition of ``Submit equipment rental request'' and ``Select suitable equipment''.

\begin{table}[th!]
\footnotesize
\caption{Structured Prompt Components for Value-added Analysis. The optimal choice for each component is highlighted in bold.}
\label{tab:value_adding}
\begin{tabularx}{\columnwidth}{lX}
\toprule
\textbf{Component} & \textbf{Variations} \\
\midrule[\heavyrulewidth]
Role Description& Neutral Analyst (Baseline), \textbf{LEAN Analyst (Expert)}, Business Consultant, Process Engineer, Customer Advocate, SME (per sector), \textbf{SME (Detailed)}, Quality Assurance Specialist \\
\midrule[\lightrulewidth]
Task Description & \textbf{Standard Classification}, Efficiency-Focused Classification, Waste Identification \\
\midrule[\lightrulewidth]
Guidelines & Basic Guidelines, Context-Aware Guidelines, \textbf{Lean Principles Guidelines} \\
\midrule[\lightrulewidth]
Classification Types & Basic, \textbf{Detailed}, Textbook, Contextualised \\
\midrule[\lightrulewidth]
Example Outputs & \textbf{Simple Process Example}, Complex Process Example, Varied Process Examples \\
\midrule[\lightrulewidth]
Context & Focus on Customer Value, Consider Regulatory Requirements, \textbf{Include Justifications} \\
\bottomrule
\end{tabularx}
\end{table}

As for the activity breakdown, we devise a {\it zero-shot baseline} that asked the LLM to classify the substeps without additional context or detailed instructions. We then again devise a range of {\it Structured Prompts} specific to the value-added analysis based on domain knowledge and best-practices in prompting. The components are summarized in Table~\ref{tab:value_adding}. We used the same grid-search methodology as for activity breakdown to identify the best-performing combination of component values (bold elements in the table). Figure~\ref{fig:prompt_vaa} presents the final best-performing prompt.

%% file: tex/evaluation.tex
\section{Evaluation}
\label{subsec:experimental_setup}

This section presents the evaluation of our approach, which includes the zero-shot baseline and the structured prompt models. These configurations are tested against a ground-truth dataset. The task breakdown and value-added classification were evaluated separately.
The rest of the section is organised as follows. We start by presenting the dataset used and our experimental setup in Section~\ref{subsec:dataset}. Then, the evaluation for the activity breakdown and value-added analysis are presented in Sections~\ref{subsec:activity_breakdown_results} and~\ref{subsec:value_added_results}, respectively.

\subsection{Dataset and experimental setup}
\label{subsec:dataset}

To evaluate our framework, we curated a dataset of 50 publicly available business process models. These models were sourced from the textbook "Fundamentals of Business Process Management"~\cite{dumas2018fundamentals} and the SAP Signavio Academic Models (SAP-SAM) repository~\cite{sola2022sap}. The processes span various domains, including banking, customer service, healthcare, manufacturing, and technology.
These models were annotated with the activity breakdown and the value-added classification for each step.
The process models were randomly split into a development set (33 models) used for model and prompt engineering, and evaluation set (17 models).

\begin{figure}
    \centering
    \includegraphics[width=\linewidth]{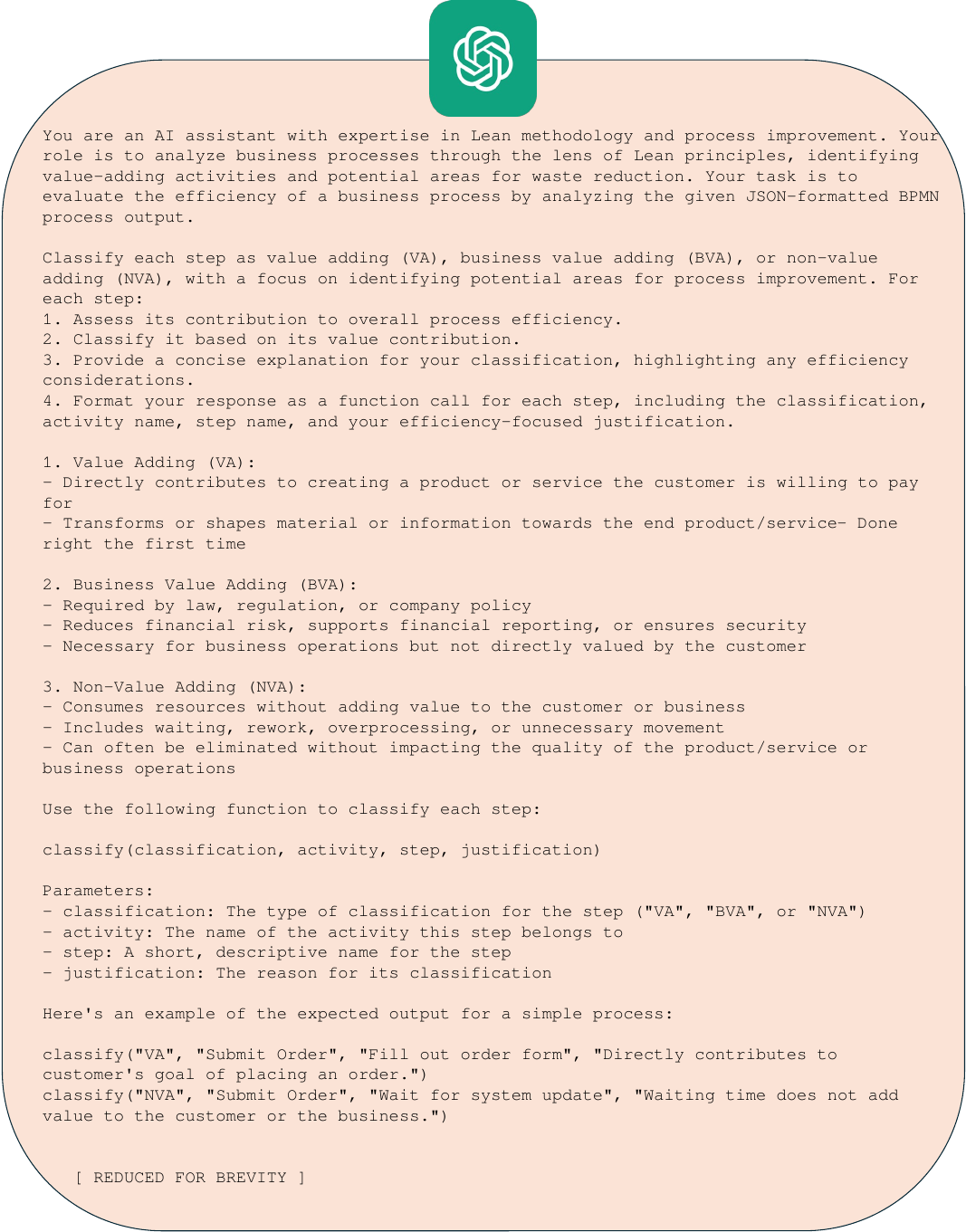}
    \caption{The final prompt for value-adding analysis.}
    \label{fig:prompt_vaa}
\end{figure}
\paragraph{Experimental setup} All experiments were done using GPT-3.5-Turbo-0125~\footnote{GPT-3.5 Turbo: \url{https://platform.openai.com/docs/models/gpt-3-5-turbo}, Accessed: 2024-10-25}. We set the model temperature to 0.1 to reduce randomness and ensure replicability~\cite{peeperkorn2024temperaturecreativityparameterlarge}. A context window of 4096 tokens was sufficient for our process descriptions and analysis tasks. All other model parameters were kept at their default values.

\subsection{Activity Breakdown}
\label{subsec:activity_breakdown_results}

The evaluation of activity breakdown is challenging because there are many valid breakdowns with different terminology or granularity.
In fact, to evaluate the subjective nature of this operation, four annotators with basic background on BPM knowledge were engaged and independently annotated a subset of models. Annotators were provided with the process model in BPMN notation and the process domain.
All annotators had taken a BPM course and were familiar with value-added analysis. The results revealed variation in the activity breakdown, largely due to paraphrasing of equivalent steps, and the variation in the granularity of the steps. Some variation was due to different customer's value perception. These variations underscore the complex nature of value assessment in business processes and the role of individual expertise and perspective in such evaluations.

Considering the variation in acceptable breakdown solutions, traditional string matching or semantic similarity metrics are insufficient for a meaningful evaluation of model-generated activity breakdown. To address this challenge, we developed a \emph{Comparator LLM}, an LLM-based framework to evaluate step alignments mirroring human expert judgment.
The comparator identifies matches between ground truth and the generated steps, considering various valid forms of alignment: \textit{exact matches} (where steps are identical), \textit{semantic matches} (paraphrases of the same underlying action), \textit{granularity variations} (where steps at different levels of detail), and {\it no match} (steps with no pendant in the ground truth, indicating an error). The comparator LLM was evaluated through manual validation on 20\% of the data set, where its classifications were found to be closely aligned with human expert judgment. We report the \% of model-generated steps for each of the four possible forms of alignment.

After applying a greedy grid-search (see Sec.~\ref{subsec:activity_breakdown}) to identify the optimal prompt components using our 33 development models, we evaluated the zero-shot baseline, and the best-performing structured prompts on the test set of 17 process models.
Table~\ref{tab:activity_breakdown_match_types_test} shows the results of the evaluation of the zero-shot baseline and the three best performing role descriptions. For BPE, SME and PM (Detailed), we used the best performing components: Breakdown with Dependencies, Detailed Guidelines, Process-Focused, Few-Shot (multiple examples) and Include Business Context (see Table~\ref{tab:activity_breakdown}).
Note that for ``Exact Match'' and ``Functional Equivalence'' high scores are best, while for ``No match'' low scores are best. Most structured models outperform the baseline across the board. The Business Process Expert (BPE) role leads to best performance, overall.

\begin{table*}[th!]
\begin{tabular}{lcP{2.5cm}P{2.5cm}c}
\toprule
              & Exact Match & Functional Equivalence & Granularity Difference & No Match \\\midrule
{Zero-shot Baseline}      & 13.4 & 29.0 & 21.3 & 36.3 \\
BPE           & \textbf{20.8} & \textbf{38.9} & \textbf{10.1} & \textbf{28.2}\\
SME           & 11.6 & 37.4 & 19.8 & 31.0\\
PM (Detailed) & 6.9 & 35.9 & 20.2 & 37.0 \\\bottomrule
\end{tabular}
\caption{Activity breakdown performance for the zero-shot baseline, and three versions of the structured prompt with different role types and other components (see bold values in Table~\ref{tab:activity_breakdown}). BPE=Business Process Expert, SME = Subject Matter Expert, PM = Project Manager.}
\label{tab:activity_breakdown_match_types_test}
\end{table*}

In the case of BPE, the alignment between LLM-generated steps and expert-provided ground truth, with almost 60\% of steps being exactly or functionally equivalent, demonstrates the LLM's capability to decompose activities effectively. The modular prompt engineering approach appears successful in guiding the LLM to produce outputs that are consistent with expert analyses.
The presence of granularity differences in 10.1\% of steps reflects the inherent subjectivity in determining the appropriate level of detail.
The 28.2\% of steps classified as ``No Match'' highlight areas where the LLM's understanding diverges from expert interpretations. While these discrepancies may stem from contextual misinterpretations or insufficient domain-specific knowledge within the LLM, these are typically expected given the subjective nature of activity breakdowns.

While these results are promising, we envision our approach as a support tool to be used as a first point of generation by an analyst. The automatically generated breakdown would later be revised; thus, the potential for misalignment is low impact, as the analyst would be expected to update these breakdowns. Please note that due to space restrictions, additional results and details about the Comparator LLM can be found in the GitHub repository.

\subsection{Value-Added Analysis}
\label{subsec:value_added_results}

This subsection presents the LLM's ability to correctly classify process steps as Value-Adding (VA), Business Value-Adding (BVA), or Non-Value-Adding (NVA).
Similarly to the activity breakdown, we asked the four human annotators to classify a given set of steps for a process. We assessed the inter-annotator agreement using Krippendorf's $\alpha$~\cite{krippendorff2011computing}, a widely used statistical measure of agreement across multiple items and annotators. The inter-annotator agreement was 0.53, indicating moderate agreement.

When using the LLMs to classify the steps in the test set, the results show 45\% (N=929) steps were annotated as VA, 48\% (N=992) were annotated as BVA and 6.3\% (N=130) as NVA.
The relatively even distribution between VA and BVA steps, coupled with the small proportion of NVA steps, aligns with what we would expect in real-world business processes that have undergone some degree of optimisation but may still contain hidden inefficiencies. Such inefficiencies would be more evident when the analysts have deeper knowledge about the process, e.g., by providing information about the handovers -- a common NVA step in business processes.

\begin{table}[th!]
\centering
\begin{tabular}{lP{2cm}P{2cm}}
\toprule
                        & F1 (Overall) & F1 (NVA) \\\midrule
{Zero-shot baseline}& 0.53 & 0.23\\
BC           & 0.66 & 0.28\\
CA               & 0.60 & 0.49\\
SME (Detailed)          & \textbf{0.72} & 0.20\\
LEAN Analyst (Expert) & 0.61 & \textbf{0.50}\\\bottomrule
\end{tabular}
    \caption{Value-added analysis performance for the zero-shot baseline (top) and four versions of the structured prompt with the best-performing role-types (rows) and other component values set (see bold values in Table~\ref{tab:value_adding}). BC=Business Consultant; CA=Customer Advocate; SME=Subject Matter Expert; LEAN Analyst (Detailed). }
    \label{tab:test_metrics_comparison}
\end{table}

The classification generated by the LLM was directly compared against the human labels in our ground truth data. The overall performance is reported as {\it F1 Score (Overall)}. In practice, businesses are particularly interested in identifying NVA steps. For that reason, we separately report performance on this class only (\textit{NVA F1-Score}). For a better understanding of patterns of model misclassification, we also include a {\it confusion matrix analysis}.

The main results are shown in Table~\ref{tab:test_metrics_comparison}. All structured prompt models outperform the zero-shot baseline (except for SME on NVA). The SEM (Detailed) configuration achieved the highest overall macro F1 score. However, the LEAN Analyst (Expert) configuration demonstrated superior waste identification capabilities with the best F1 NVA score, while retaining an acceptable overall F1 score. 
The LEAN Analyst (Expert) shows effective identification of NVA steps and demonstrates its potential as a tool for waste identification when appropriately guided.

\begin{wrapfigure}{r}{0.55\textwidth}
    \vspace{-4mm}
    \includegraphics[width=0.98\linewidth,clip,trim=0 0 0.8cm 16cm]{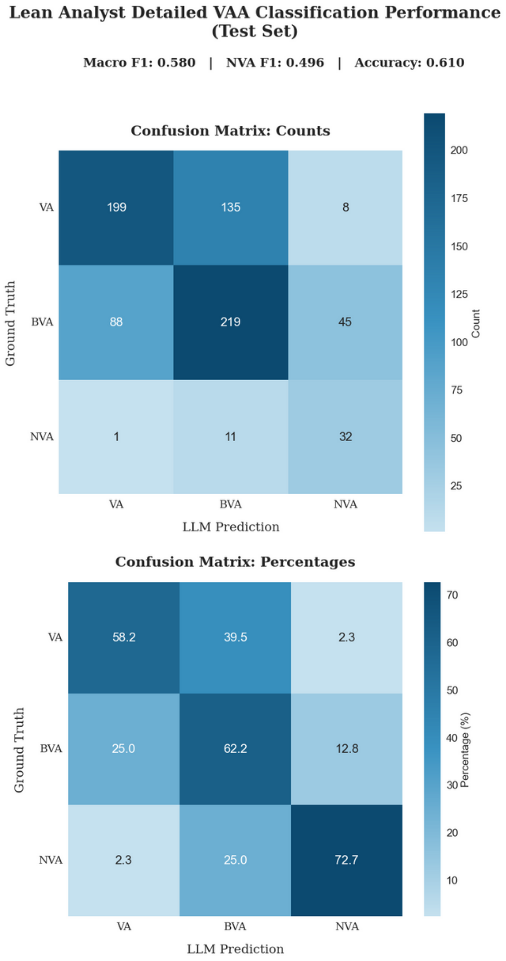}
    \caption{Confusion matrix for the LEAN Expert (Detailed) configuration on the test set.}
    \label{fig:test_confusion_matrix}
        \vspace{-2mm}
\end{wrapfigure}
We next investigate the patterns of misclassification by the LEAN Analyst.
The confusion matrix in Fig.~\ref{fig:test_confusion_matrix} shows the comparison between the LLM and the ground truth. In the matrix, 72.7\% of true NVA steps were correctly classified and only 2.3\% misclassified as VA. The model showed some difficulty distinguishing between VA and BVA categories, with 39.5\% of VA activities classified as BVA, reflecting the subtle distinctions between customer value and business necessity.

\section{Discussion}
\label{sec:discussion}
Our evaluation results showed the potential of our framework for automated value-added analysis, highlighting both its strengths and areas for improvement. Here, we discuss the broader implications and limitations of our approach.

\subsection{Implications}
\label{subsec:implications_limitations}

The integration of LLMs into business process analysis through our framework demonstrates potential for transforming traditional BPM practices, particularly in standardising and scaling process evaluation.

The framework's ability to automate complex tasks traditionally performed manually enables organisations to conduct more frequent and comprehensive process evaluations at scale. By leveraging standardised guidelines and prompt structures, our framework promotes consistency in analysis outputs—particularly valuable in large organisations where multiple analysts may be involved in process improvement efforts. Furthermore, LLMs' advanced language understanding capabilities can potentially uncover insights that might be overlooked by human analysts, enhancing the depth of process optimisation.

Our framework may be most effectively implemented as an interactive tool rather than a replacement for human expertise. The framework excels at standardising routine analyses and identifying potential areas of waste, but human oversight remains crucial for contextualising results and making strategic improvement decisions. This hybrid approach leverages our framework's consistency and scalability while maintaining the nuanced judgement necessary for effective process improvement.
\
\subsection{Limitations and Challenges}
\label{subsec:limitations_challenges}

Despite its strengths, our framework faces several limitations that warrant careful consideration:

\begin{itemize}
    \item \textbf{Subjectivity and Context Dependence.} The moderate inter-annotator agreement among experts (Krippendorff's $\alpha=0.53$ for the value-added analysis) highlights the subjective nature of activity decomposition and value classification. This subjectivity extends to the LLM, which may produce varying outputs based on prompt phrasing and context. Establishing more robust guidelines and incorporating context-aware mechanisms could improve consistency.
    \item \textbf{Prompt Sensitivity and Design Complexity.} The LLM's sensitivity to prompt variations necessitates careful prompt engineering, which can be resource-intensive. The greedy grid search approach, while effective, may not capture all nuances influencing the LLM's performance. Developing systematic methods for prompt optimisation and exploring adaptive prompting strategies could enhance reliability.
    \item \textbf{Data and Domain Dependence.} Our prompt selection outcome depended on the 33 process models used in the process, which may not encompass all industry-specific terminology or practices. This limitation can lead to misinterpretations or omissions in specialised and nuanced domains. Integrating domain-specific corpora into the LLM's training data or employing fine-tuning techniques could improve domain relevance.
    \item \textbf{Transparency and Explainability.} While LLMs can provide justifications for their classifications, these explanations may lack the depth or clarity required for critical business decisions. Enhancing the explainability of the LLM's reasoning process is essential for building trust and facilitating acceptance among stakeholders.
\end{itemize}

\subsection{Ethical and Privacy Considerations}
\label{subsec:ethical_considerations}

The application of LLMs to business process analysis raises important privacy concerns, particularly regarding the processing of sensitive organisational data. Companies implementing such systems must ensure robust data protection measures and carefully consider the implications of processing proprietary process information through external AI systems. Of particular concern is the potential for process information to be exposed through model retraining, necessitating clear data handling protocols and privacy guarantees.

%% file: tex/conclusion.tex
\section{Conclusion and Future Work}
\label{sec:conclusion}

This paper introduced a novel approach using Large Language Models (LLMs) for automated value-added analysis. This occurs in two stages:
\begin{inparaenum}[]
    \item[] First, activities of a process are divided into steps using standardised guidelines and modular prompt engineering, facilitating granular analysis.
    \item[] Second, each step is classified into Value Adding (VA), Business Value Adding (BVA), and Non-Value Adding (NVA) categories, enabling targeted waste identification.
\end{inparaenum} 
The experimental results demonstrate that our framework effectively automates complex tasks with a reasonable degree of accuracy and identification of waste. The use of a greedy grid search strategy for prompt optimisation proved instrumental in enhancing the LLM's performance, underscoring the importance of prompt design in AI-driven analysis.

The presented approach represents an advancement in the application of AI to business process management. It demonstrates the feasibility and benefits of integrating LLMs into process analysis and waste identification efforts. While the approach was presented as fully automated, we recognize that the integration of human expertise through a human-in-the-loop approach is necessary to obtain results that are more relevant for the analysed process.


There are various directions for future work. 
\begin{inparaenum}[]
    \item The framework's sensitivity to prompt design, as well as the fast-moving field of LLMs themselves, necessitates ongoing development of prompting strategies. Future research should explore reinforcement learning for automated prompt tuning \cite{lester2021power} and develop meta-prompting \cite{suzgun2024meta} approaches for context-aware prompting that adjust to specific process characteristics.
    \item Another direction of future work is to leverage external information to improve the prompting and analysis; such as the information captured in business process event logs (e.g., resources or timestamps). By incorporating event logs and operational data, future versions could provide data-driven validation of LLM outputs and extend into predictive analysis to identify potential process inefficiencies.
    \item Finally, another direction for future work is to extend the capabilities of our approach to automated process redesign~\cite{malhotra1998business}. This development would transform our framework from an analytical tool into a comprehensive process improvement platform, capable of not only identifying inefficiencies but also suggesting and evaluating process improvements. 
\end{inparaenum}